\documentclass[sigconf]{acmart}\usepackage{color}

\usepackage{multirow}
\usepackage{pifont}

\AtBeginDocument{%
  \providecommand\BibTeX{{%
    \normalfont B\kern-0.5em{\scshape i\kern-0.25em b}\kern-0.8em\TeX}}}

\setcopyright{acmcopyright}
\copyrightyear{2022}
\acmYear{2022}
\acmConference[MM'22]{Proceedings of the 30th ACM International Conference on Multimedia}{October. 10--14, 2022}{Lisbon, Portugal}
\acmBooktitle{Proceedings of the 30th ACM International Conference on Multimedia (MM'22) October. 10--14, 2022, Lisbon, Portugal}
\acmPrice{15.00}
\acmISBN{xxxxxx}
\acmDOI{10.1185/xxxxxx}

\begin{document}

\title{Magic ELF: Image D\underline{e}raining Meets Association \underline{L}earning and Trans\underline{f}ormer}

% \author{Kui Jiang$^1$},
% \author{Zhongyuan Wang$^{1}$}
% \authornote{Corresponding Author.}
% \author{Chen Chen$^2$}
% \author{Zheng Wang$^{1*}$},
% \authornotemark[1]
% \author{Laizhong Cui$^3$}
% \author{Chia-Wen Lin$^4$}
% %\authornote{Equal Contribution.}
% \affiliation{\normalsize{${^1}$
% \institution{NERCMS, School of Computer Science, Wuhan University, Wuhan 430072, China }}
% }
% \affiliation{\normalsize{${^2}$
% \institution{Center for Research in Computer Vision, University of Central Florida}}
% }
% \affiliation{\normalsize{${^3}$
% \institution{College of Computer Science and Software Engineering, Shenzhen University}}
% }
% \affiliation{\normalsize{${^4}$
% \institution{National Tsing Hua University}}
% }
%\email{{shunmh,kuijiang,niezhixiang,wangzwhu}@whu.edu.cn}

\author{Kui Jiang}
\affiliation{%
 \institution{NERCMS, Wuhan University \country{China}}
}

\author{Zhongyuan Wang}\authornote{Corresponding author}
\affiliation{%
 \institution{NERCMS, Wuhan University \country{China}}
}

\author{Chen Chen}
\affiliation{%
%  \institution{Center for Research in Computer Vision, University of Central Florida \country{U.S.}}
  \institution{CRCV, University of Central Florida \country{U.S.}}
}

\author{Zheng Wang}\authornotemark[1]
\affiliation{%
 \institution{NERCMS, Wuhan University \country{China}}
}

\author{Laizhong Cui}
\affiliation{%
%  \institution{College of Computer Science and Software Engineering, Shenzhen University \country{China}}
% }
 \institution{Shenzhen University \country{China}}
}

\author{Chia-Wen Lin}
\affiliation{%
 \institution{National Tsing Hua University  \country{China}}
}
%\thanks
%\renewcommand{\shortauthors}{Jiang and Wang, et al.}
\renewcommand{\shortauthors}{Kui Jiang et al.}

% \author{Kui Jiang$^{1}$ \quad Zhongyuan Wang$^{1,\dagger}$ \quad Chen Chen$^{2}$ \quad Zheng Wang$^{1,\dagger}$ \\Laizhong Cui$^{3}$ \quad Chia-Wen Lin$^{4}$}
% \affiliation{%
%  \institution{$^1$National Engineering Research Center for Multimedia Software, School of Computer Science, Wuhan University \country{China}}
%  \institution{$^2$Center for Research in Computer Vision, University of Central Florida \country{U.S.}}
%   \institution{$^3$College of Computer Science and Software Engineering, Shenzhen University \country{China}}
%   \institution{$^4$National Tsing Hua University \country{Taiwan}}
% }
% \thanks{$^\dagger$Corresponding author.}

% \renewcommand{\shortauthors}{Jiang and Wang, et al.}

%%
%% The abstract is a short summary of the work to be presented in the
%% article.
\begin{abstract}
Convolutional neural network (CNN) and Transformer have achieved great success in multimedia applications. However, little effort has been made to effectively and efficiently harmonize these two architectures to satisfy image deraining. This paper aims to unify these two architectures to take advantage of their learning merits for image deraining. In particular, the local connectivity and translation equivariance of CNN and the global aggregation ability of self-attention (SA) in Transformer are fully exploited for specific local context and global structure representations. Based on the observation that rain distribution reveals the degradation location and degree, we introduce degradation prior to help background recovery and accordingly present the association refinement deraining scheme. A novel multi-input attention module (MAM) is proposed to associate rain perturbation removal and background recovery. Moreover, we equip our model with effective depth-wise separable convolutions to learn the specific feature representations and trade off computational complexity. Extensive experiments show that our proposed method (dubbed as ELF) outperforms the state-of-the-art approach (MPRNet) by 0.25 dB on average, but only accounts for 11.7\% and 42.1\% of its computational cost and parameters. The source code is available at \url{https://github.com/kuijiang94/Magic-ELF }.
\end{abstract}

\begin{CCSXML}
<ccs2012>
   <concept>
       <concept_id>10010147.10010178.10010224</concept_id>
       <concept_desc>Computing methodologies~Computer vision</concept_desc>
       <concept_significance>500</concept_significance>
       </concept>
   <concept>
       <concept_id>10010147.10010257.10010321</concept_id>
       <concept_desc>Computing methodologies~Machine learning algorithms</concept_desc>
       <concept_significance>300</concept_significance>
       </concept>
 </ccs2012>
\end{CCSXML}
\ccsdesc[500]{Computing methodologies~Computer vision}
\ccsdesc[300]{Computing methodologies~Machine learning algorithms}

\keywords{Image Deraining, Self-attention, Association Learning}

\maketitle
\section{Introduction}

\begin{figure}[!ht]
\flushleft
\centering
\includegraphics[width=0.48\textwidth]{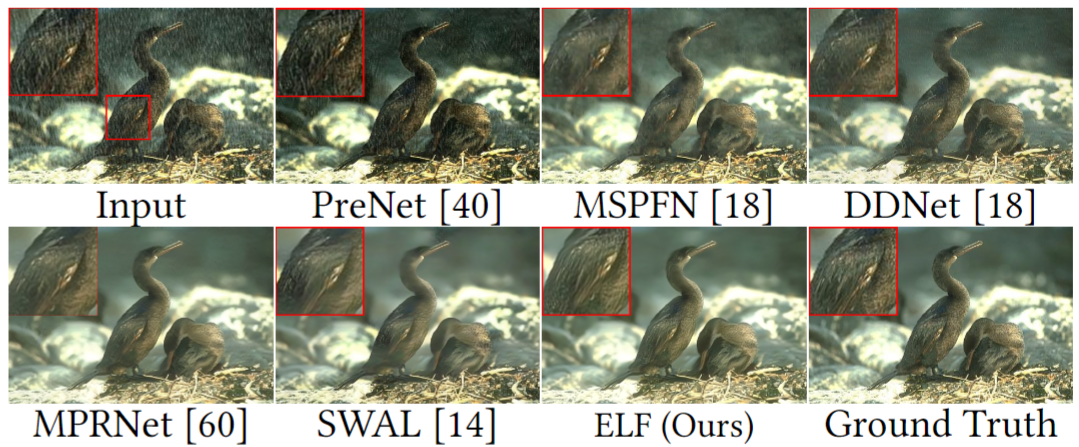}
%\vspace*{-4mm}%%\vspace*{-4mm}
\caption{Visual comparison results. PreNet~\cite{ren2019progressive} and DRDNet~\cite{9157472} remove rain streaks in some cases, but fail to cope with additional degradation effects of \textbf{missing details and contrast bias}. MPRNet~\cite{zamir2021multi} and SWAL~\cite{huang2021selective} tend to smooth the contents while our ELF reconstructs credible textures. Please refer to the region highlighted in the \textcolor{red}{red} boxes for a close up comparison.}
%embedding representation of $I^*_{B,S}$.}
% which is helpful for the accurate texture recovery of reconstruction sub-network (RSN).} %for texture reconstruction and resolution amplification.}
\label{fig:sample}
\vspace{-2mm}
\end{figure}

Rain perturbation causes detrimental effects on image quality and significantly degrades the performance of multimedia applications like image understanding~\cite{liao2022unsupervised,wang2022med},
object detection~\cite{zhong2021unsupervised} and identification~\cite{xie2022sampling}.
Image deraining~\cite{chen2022unpaired,wang2020dcsfn,jiang2021multi} tends to produce the rain-free result from the rainy input, and has drawn widespread attention in the last decade. Prior to deep neural networks, the early model-based deraining methods~\cite{garg2005does} rely more on statistical analyses of image contents, and enforce handcrafted priors (\textit{e.g.}, sparsity and non-local means filtering) on both rain and background. Still, they are not robust to varying and complex rain conditions~\cite{bossu2011rain, chen2013generalized, Zhong2022RainyWA}.

Because of the powerful ability to learn generalizable priors from large-scale data, CNNs have emerged as a preferable choice compared to conventional model-based methods. To further promote the deraining performance, various sophisticated architectures and training practices are designed to boost the efficiency and generalization~\cite{9294056,yang2021end,yu2019gradual,yang2021end}. However, due to intrinsic characteristics of local connectivity and translation equivariance, CNNs have at least two shortcomings: 1) limited receptive field; 2) static weight of sliding window at inference, unable to cope with the content diversity. The former thus prevents the network from capturing the long-range pixel dependencies while the latter sacrifices the adaptability to the input contents. As a result, it is far from meeting the requirement in modeling the global rain distribution, and generates results with obvious rain residue (PreNet~\cite{ren2019progressive} and DRDNet~\cite{9157472}) or detail loss (MPRNet~\cite{zamir2021multi} and SWAL~\cite{huang2021selective}). Please refer to the deraining results in Figure~\ref{fig:sample}.
%especially for the globally distributed rain streaks.
%While several variants and strategies, including the dilated convolution, deformation convolution and U-shaped structure, have been developed to cover larger fields and select the most related components for better long-range correlation learning, the marginal improvement comes at extra optimization burden and computation consumption. In particular, the learnable correlation mappings are still sparse, \textit{e.g.}, only 9 positions for a $3\times 3$ kernel. As a result, it is far from meeting the requirement in modeling the global correlation, especially for the globally distributed rain streaks.

\begin{figure}
\centering
		\begin{tabular}{cc}
		\includegraphics[width=0.48\columnwidth]{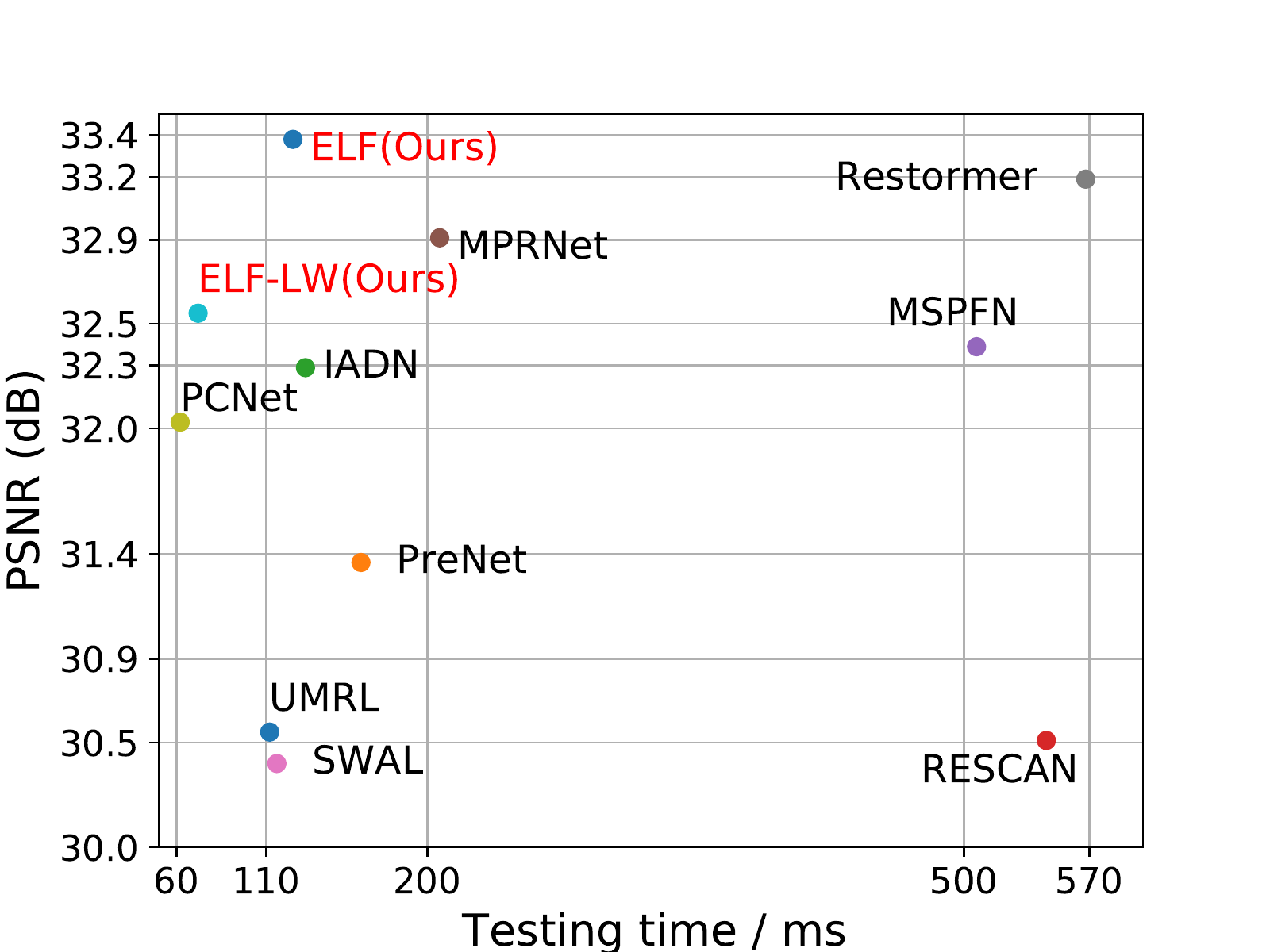}&\includegraphics[width=0.48\columnwidth]{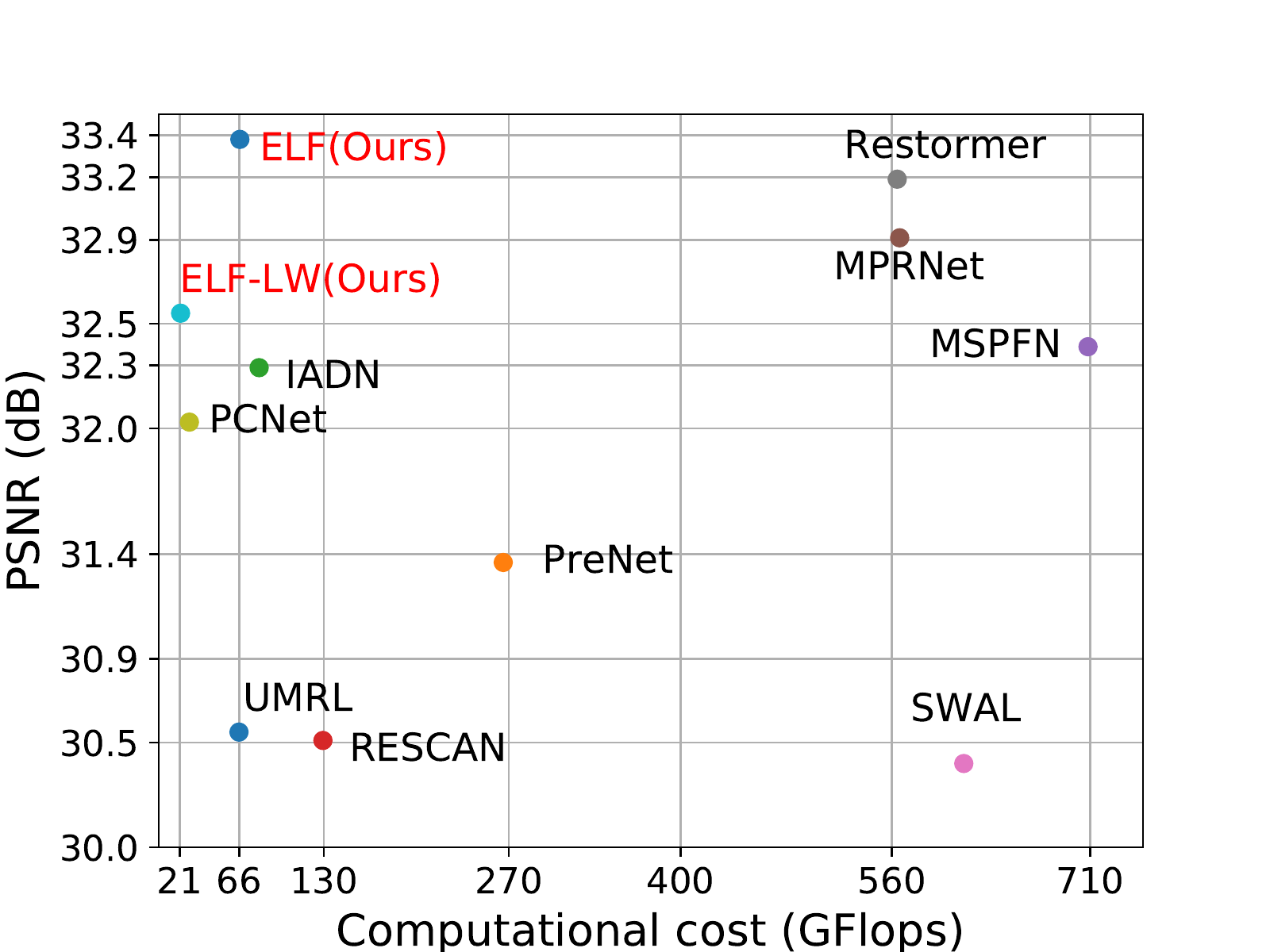}\\
% 			\includegraphics[width=0.9\columnwidth]{./img/psnrtime.pdf}\\
% 			\includegraphics[width=0.9\columnwidth]{./img/psnrgflops.pdf} \\
            %\tiny{Real-world Sample} & \tiny{Synthetic Sample}  \\
	\end{tabular}\vspace{-2mm}
\caption{Comparison of mainstream deraining methods in terms of efficiency (inference time (ms) and computational cost (Gflops)) \textit{vs.} performance (PSNR) on the \textit{TEST1200} dataset with image size of \textit{$512\times 512$}. Compared with the top-performing method Restormer~\cite{9157472}, our ELF achieves comparable deraining performance (\textit{33.38dB \textit{vs.} 33.19dB}) while saving 88.0\% inference time (ms) (\textit{125 \textit{vs.} 568}) and 88.4\% computational cost (Gflops) (\textit{66.39 \textit{vs.} 568}). Our light-weight model ELF-LW is still competitive, surpassing the real-time deraining method PCNet by 0.52dB while with less computation cost  (Gflops) (\textit{21.53 \textit{vs.} 28.21}). }
%Our light-weight model ELF-LW not only achieves real-time throughput (\textit{35.7fps}) but also outperforms the representative high-accuracy method PreNet~\cite{ren2019progressive}.}
% RESCAN~\cite{li2018recurrent} and UMRL~\cite{yasarla2019uncertainty})}
\vspace{-2em}
\label{fig:example}
\end{figure}

%cover larger fields and select the most related components for better long-range correlation learning
Self-attention (SA) calculates response at a given pixel by a weighted sum of all other positions, and thus has been explored in deep networks for various natural language and computer vision tasks~\cite{vaswani2017attention,wang2018non,zhang2019self}.
%Multi-head self-attention, as a unique implementation, is optimized for parallelization and effective representation learning, which is widely spread and has shown impressive performance~\cite{brown2020language,dosovitskiy2020image,liang2021swinir}.
Benefiting from the advantage of global processing, SA achieves significant performance boost over CNNs in eliminating the degradation perturbation~\cite{liang2021swinir, chen2021pre, wang2021uformer}.
%its powerful representation ability in capturing long-range pixel relations, SA has also been employed in image restoration to eliminate the degradation perturbation~\cite{liang2021swinir, chen2021pre, wang2021uformer}.
However, due to the global calculation of SA, its computation complexity grows quadratically with the spatial resolution, making it infeasible to apply to high-resolution images~\cite{zamir2021restormer}. More recently, Restormer~\cite{zamir2021restormer}  proposes a multi-Dconv head ``transposed'' attention (MDTA) block to model global connectivity, and achieves impressive deraining performance. Although MDTA applies SA across feature dimension rather than the spatial dimension and has linear complexity, still, Restormer quickly overtaxes the computing resources.
%It has 563.96 GFlops and 26.10 Million parameters, and consumes 0.568s to derain a single image with $512\times512$ pixels using one TITAN X GPU), which is computationally or memory expensive for many real applications that require online processing of data with low latency on resource-constrained devices.
As illustrated in Figure~\ref{fig:example}, the high-accuracy model Restormer~\cite{zamir2021restormer} requires much more computation resource for a better restoration performance. It has 563.96 GFlops and 26.10 Million parameters, and consumes 0.568s to derain an image with $512\times512$ pixels using one TITAN X GPU), which is computationally or memory expensive for many real-world applications with resource-constrained devices.
%that require online processing of data with low latency on resource-constrained devices.
%the light-weight deraining model LPNet~\cite{fu2019lightweight} leads to unsatisfactory deraining performance, whereas the high-accuracy models RESCAN~\cite{li2018recurrent} and MSPFN~\cite{9157472} require much more computation resource for a better restoration performance.

Besides low efficiency, there are at least another two shortcomings for Restormer~\cite{zamir2021restormer}. 1) Regarding the image deraining as a simple rain streaks removal problem based on the additive model is debatable, since the rain streak layer and background layer are highly interwoven, where rain streaks destroy image contents, including the details, color, and contrast. 2) Constructing a pure Transformer-based framework is suboptimal, since SA is good at aggregating global feature maps but immature in learning local contexture relations which CNNs are skilled at. That in turn naturally raises two questions: \textit{(1) How to associate the rain perturbation removal and background recovery? (2) How to unify SA and CNNs efficiently for image deraining?}
%employ the SA work for image deraining tasks in a more efficient manner? %To this end, an elaborate deraining model to unify the respective learning merits of CNNs and SA, while with high accuracy and efficiency is urgently required.

To answer the first question, we take inspiration from the observation that rain distribution reflects the degradation location and degree, in addition to the rain distribution prediction. Therefore, we propose to refine
%To solve the first issue, based on the observation that rain distribution reflects the degradation location and degree, in addition to the rain distribution prediction, we propose to refine
background textures with the predicted degradation prior in an association learning manner. As a result, we accomplish the image deraining by associating rain streak removal and background recovery, where an image deraining network (IDN) and a background recovery network (BRN) are specifically designed for these two subtasks. The key part of association learning is a novel multi-input attention module (MAM). It generates the degradation prior and produces the degradation mask according to the predicted rainy distribution. Benefited from the global correlation calculation of SA, MAM can extract the informative complementary components from the rainy input (query) with the degradation mask (key), and then help accurate texture restoration.
%To achieve the optimal approximation, we introduce joint constraints for enhancing the compatibility of the deraining model with background recovery, automatically learned from the training data.

%It has been demonstrated that the SA and standard convolution exhibit opposite behaviors but complementary~\cite{park2022vision}. Specifically, SA tends to aggregate feature maps with self-attention importance, but convolution diversifies them to focus on the local textures. %In other words, SA is more like low-pass filters, but convolutions are high-pass filters.
An intuitive idea to deal with the second issue is to construct a unified model with the advantages of these two architectures.
It has been demonstrated that the SA and standard convolution exhibit opposite behaviors but complementary~\cite{park2022vision}. Specifically, SA tends to aggregate feature maps with self-attention importance, but convolution diversifies them to focus on the local textures. Unlike Restormer equipped with pure Transformer blocks, we promote the design paradigm in a parallel manner of SA and CNNs, and propose a hybrid fusion network. It involves one residual Transformer branch (RTB) and one encoder-decoder branch (EDB) (The detailed pipeline
is provided in \textcolor{blue}{Supplementary}.). The former takes a few learnable tokens (feature channels) as input and stacks multi-head attention and feed-forward networks to encode global features of the image. The latter, conversely, leverages the multi-scale encoder-decoder to represent contexture knowledge.
We propose a light-weight hybrid fusion block (HFB) to aggregate the outcomes of RTB and EDB to yield a final solution to the subtask.
%It leverages the efficient depthwise and pointwise convolution to extract local features. Former takes a few learnable tokens as input and stacks multi-head attention and feed-forward networks (FFN). These tokens are used to encode global features of the image
%we propose a two-path hybrid fusion network, involving one residual Transformer branch (RTB) and one encoder-decoder branch (EDB) (\textit{The detailed pipeline is provided in \textcolor{blue}{Supplementary}.}).
%The former is used to model the global structure through the adaptive correlation learning (low-frequency components) while the latter, conversely, predicts the contexture knowledge (high-frequency components). The outcomes of these two branches can be combined to yield a final solution to the subtask.
In this way, we construct our final model as a two-stage Transformer-based method, namely ELF, for single image deraining, which \textcolor{blue}{outperforms the CNN-based SOTA (MPRNet~\cite{zamir2021multi}) by 0.25dB on average, but saves 88.3\% and 57.9\% computational cost and parameters.} %This showcases an efficient and effective implementation of part-whole hierarchy.

The main contributions of this paper are summarized as follows.
\begin{itemize}
\item %We investigate the image deraining task, CNNs and Transformer, and harmonize Transformer and CNNs to construct an effective yet efficient association learning-based network, dubbed as ELF for single image deraining. It consists of an image deraining network (IDN) for rain streaks removal, and a background reconstruction network (BRN) for details recovery.
To the best of our knowledge, we are the first to consider the high efficiency and compatibility of Transformer and CNNs for the image deraining task, and unify the advantages of SA and CNNs into an association learning-based network for rain perturbation removal and background recovery. This showcases an efficient and effective implementation of part-whole hierarchy.

\item We design a novel multi-input attention module (MAM) to associate rain streaks removal and background recovery tasks elaborately. It significantly alleviates the learning burden while promoting the texture restoration.
%\item We come to a two-path hybrid fusion network to leverage the advantages of SA and CNNs that are capable of focusing on local and global interactions for feature representation.
%fully exploit  the input information and a spatial-channel skip fusion block (S2FB) for the effective multi-scale residue fusion.
%\item We endow ELF with residual attention, depth-wise separable convolutions, and the U-shaped structure to effectively encode rain/textural features. Moreover, these components make the overall framework resource-efficient for the practical deployment.
\item Comprehensive experiments on image deraining and detection tasks have verified the effectiveness and efficiency of our proposed ELF method. ELF surpasses MPRNet~\cite{zamir2021multi} by 0.25dB on average, while the latter suffers from 8.5$\times$ computational cost and 2.4$\times$ parameters.

%which is 7.5$\times$ higher efficiency and 1.4 $\times$ less parameters while gains 0.25dB improvement over MPRNet~\cite{zamir2021multi}.

%surpassing MPRNet~\cite{zamir2021multi} by 0.25dB but with only 11.7\% and 42.1\% computational cost and parameters.
%is more suitable for preserving the textural details while removing rain streaks.
\end{itemize}

\begin{figure*}[!ht]
\flushleft
\centering
\includegraphics[width=0.95\textwidth]{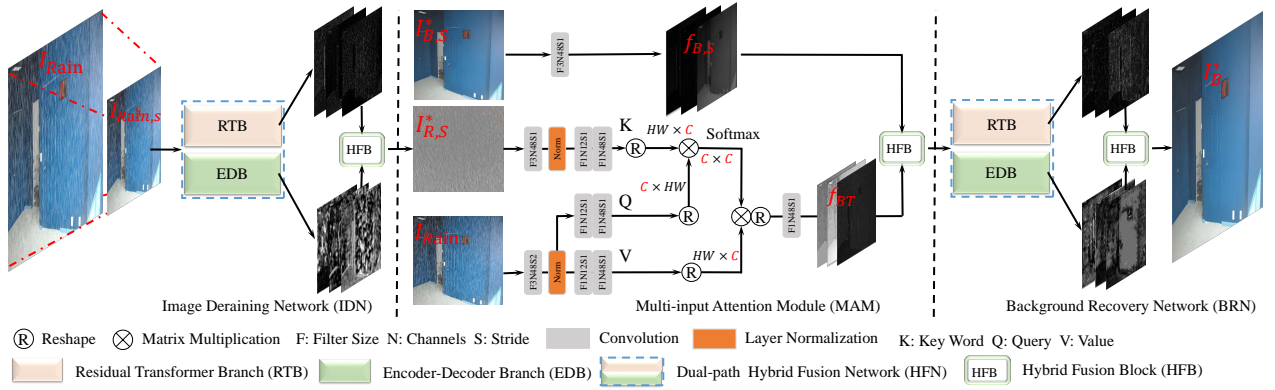}
%\vspace*{-4mm}%%\vspace*{-4mm}
\caption{The architecture of our proposed ELF deraining method. It consists of an image deraining network (IDN), a multi-input attention module (MAM), and a background reconstruction network (BRN). IDN learns the corresponding rain distribution $I^*_{R,S}$ from the sub-sample $I_{Rain,S}$, and produces the corresponding deraining result $I^*_{B,S}$ by subtracting $I^*_{R,S}$. Then, MAM takes $I^*_{R,S}$, $I_{Rain}$ and $I^*_{B,S}$ as inputs, where the predicted rain distribution provides the prior (local and degree) to exploit
complementary background components $f_{BT}$ from $I_{Rain}$ to promote the background recovery.}
%embedding representation of $I^*_{B,S}$.}
% which is helpful for the accurate texture recovery of reconstruction sub-network (RSN).} %for texture reconstruction and resolution amplification.}
\label{fig:Framework}
\vspace{-4mm}
\end{figure*}

\section{Related Work}
\label{sec:relat}
Image deraining has achieved significant progress in innovative architectures and training methods in the last few years. Next, we briefly describe the typical models for image deraining and visual Transformer relative to our studies.

\subsection{Single Image Deraining}
Traditional deraining methods~\cite{kang2012single,kang2012single,7410745} adopt image processing techniques and hand-crafted priors to address the rain removal problem. However, these methods produce unsatisfied results when the predefined model do not hold. Recently, deep-learning based approaches~\cite{li2017single,zhang2017convolutional,jiang2022multi} have emerged for rain streak removal and demonstrated impressive restoration performance. Early deep learning-based deraining approaches~\cite{fu2017removing,fu2017clearing} apply convolution neural networks (CNNs) to directly reduce the mapping range from input to output and produce rain-free results. To better represent the rain distribution, researchers take rain characteristics such as rain density~\cite{zhang2018density}, size and the veiling effect~\cite{li2017single,li2019a} into account, and use recurrent neural networks to remove rain streaks via multiple stages~\cite{li2018recurrent} or the non-local network~\cite{9157358} to exploit long-range spatial dependencies for better rain streak removal~\cite{li2018non}. Further, self-attention (SA) is recently introduced to eliminate the rain degradation with its powerful global correlation learning, and achieves impressive performance.
Although the token compressed representation and global non-overlapping window-based SA~\cite{wang2021uformer,ji2021u2} are adopted to promote the global SA to alleviate the computational burden, these models still quickly overtaxes the computing resource. Apart from the low efficiency, these methods~\cite{ji2021u2,zamir2021restormer} regard the deraining task as the rain perturbation removal only, ignoring the additional degradation effects of missing details and contrast bias.

\subsection{Vision Transformers}

Transformer-based models are first developed for sequence processing in natural language tasks~\cite{vaswani2017attention}. Due to the distinguishing feature of the strong capability to learn long-range dependencies, ViT~\cite{dosovitskiy2020image} introduces Transformer into computer vision field, and then a plenty of Transformer-based methods have been applied to vision tasks such as image recognition~\cite{dosovitskiy2020image, ijaz2022multimodal}, segmentation~\cite{wang2021pyramid}, object detection~\cite{carion2020end,liu2021swin}. Vision Transformers~\cite{dosovitskiy2020image,touvron2021training} decompose an image into a sequence of patches (local windows) and learn their mutual relationships, which is adaptable to the given input content~\cite{khan2021transformers}. Especially for low-level vision tasks, since the global feature representation promotes accurate texture inference, Transformer models have been employed to solve the low-level vision problems~\cite{liang2021swinir,wang2021uformer}.
For example, TTSR~\cite{yang2020learning} proposes a self-attention module to transfer the texture information in the reference image to the high-resolution image reconstruction, which can deliver accurate texture features. Chen~\emph{et al.}~\cite{chen2021pre} propose a pre-trained image processing transformer on the ImageNet datasets and uses the multi-head architecture to process different tasks separately. However, the direct application of SA fails to exploit the full potential of Transformer, resulting from heavy self-attention computation load and inefficient communications across different depth (scales) of layers. Moreover, little effort has been made to consider the intrinsic complementary characteristics between Transformer and CNNs to construct a compact and practical model.
%However, these studies are still the direct application of the Transformer, besides the quadratic computational complexity, and they barely consider the intrinsic complementary characteristics between Transformer and CNNs.
Naturally, this design choice restricts the context aggregation within local neighborhoods, defying the primary motivation of using self-attention over convolutions, thus not ideally suited for image-restoration tasks. In contrast, we propose to explore the bridge, and construct a hybrid model of Transformer and CNN for image deraining task.

%However, the computational complexity of SA in Transformers can increase quadratically with the number of image patches, thereby prohibiting its application to high-resolution images. Therefore, in low-level image processing applications, where high-resolution outputs need to be generated, recent methods generally employ different strategies to reduce complexity. One potential remedy is to apply self-attention within local image regions~\cite{wang2021uformer} using the Swin Transformer design~\cite{liang2021swinir}. However, this design choice restricts the context aggregation within local neighbourhood, defying the main motivation of using self-attention over convolutions, thus not ideally suited for image-restoration tasks. In contrast, we present a Transformer model that can learn long-range dependencies while remaining computationally efficient.

\section{Proposed Method}
\label{sec:Method}

Our main goal is to construct a high-efficiency and high-accuracy deraining model by taking advantage of the CNN and Transformer. Theoretically, the self-attention (SA) averages feature map values with the positive importance-weights to learn the global representation while CNNs tend to aggregate the local correlated information. Intuitively, it is reasonable to combine them to fully exploit the local and global textures. A few studies try to combine these two structures to form a hybrid framework for low-level image restoration but have failed to give full play to it. Taking the image deraining as an example, unlike the existing Transformer-based methods that directly apply Transformer blocks to replace convolutions, we consider the high efficiency and compatibility of these two structures, and construct a hybrid framework, dubbed ELF to harmonize their advantages for image deraining.
%the intrinsic characteristics of image deraining and the SA module, and construct a compact yet effective hybrid model, dubbed ELF for image deraining.
Compared to the existing deraining methods, our proposed ELF departs from them in at least two key aspects. \textit{Differences in design concepts:} unlike the additive composite model that predicts the optimal approximation $I^*_{B}$ of background image $I_{B}$ from the rainy image $I_{Rain}$, or learns the rain residue $I^*_{R}$ and subtracting it to generate $I^*_{B}$, ELF casts the image deraining task into the composition of rain streak removal and background recovery, and introduces the Transformer to associate these two parts with a newly designed multi-input attention module (MAM). \textit{Differences in composition:} since the low-frequency signals and high-frequency signals are informative to SA and convolutions~\cite{park2022vision}, a dual-path framework is naturally constructed for the specific feature representation and fusion. Specifically, the backbone of ELF contains a dual-path hybrid fusion network, involving one residual Transformer branch (RTB) and one encoder-decoder branch (EDB) to characterize global structure (low-frequency components) and local textures (high-frequency components), respectively.

Figure~\ref{fig:Framework} outlines the framework of our proposed ELF, which contains an image deraining network (IDN), a multi-input attention module (MAM), and a background recovery network (BRN). For efficiency, IDN and BRN share the same dual-path hybrid fusion network, which are elaborated in Section~\ref{sec:HFN}.
%: Hybrid Fusion Network.
%As shown in Figure~\ref{fig:Framework}, we construct a two-stage deraining network, dubbed ELF, involving a an image deraining network (IDN), a multi-input attention module (MAM), and a background recovery network (BRN). The former is used

%, and their association learning. Thus, we propose a novel two-stage association network (ELF) for image deraining, which consists of a deraining sub-network (DSN), a reconstruction sub-network (RSN) and a prior-guidance texture enhancement module (PTEM). In particular, the rain distribution predicted in DSN is employed to provide the rain priors (``where and how much") in PTEM, guiding RSN to fully exploit the background features from the rainy input, which delivers superior performance by inferring contents with better affinity as it dampens the influence of heavily degraded/corrupted information. Considering that our ultimate goal is to reconstruct the corrupted details, we propose to simplify the first learning task by predicting and characterizing the rain distribution in the sub-space via the sampled low-resolution rainy image, which is more efficient to alleviate the computational burden. Compared with DRDNet, our ELF is more efficient and can be directly applied to large-size feature maps.
\subsection{Pipeline and Model Optimization}
\label{sec:AMO}
\begin{figure}[!t]
	\centering
	\renewcommand\arraystretch{0.3}
		\begin{tabular}{cc}
			\includegraphics[width=0.45\columnwidth]{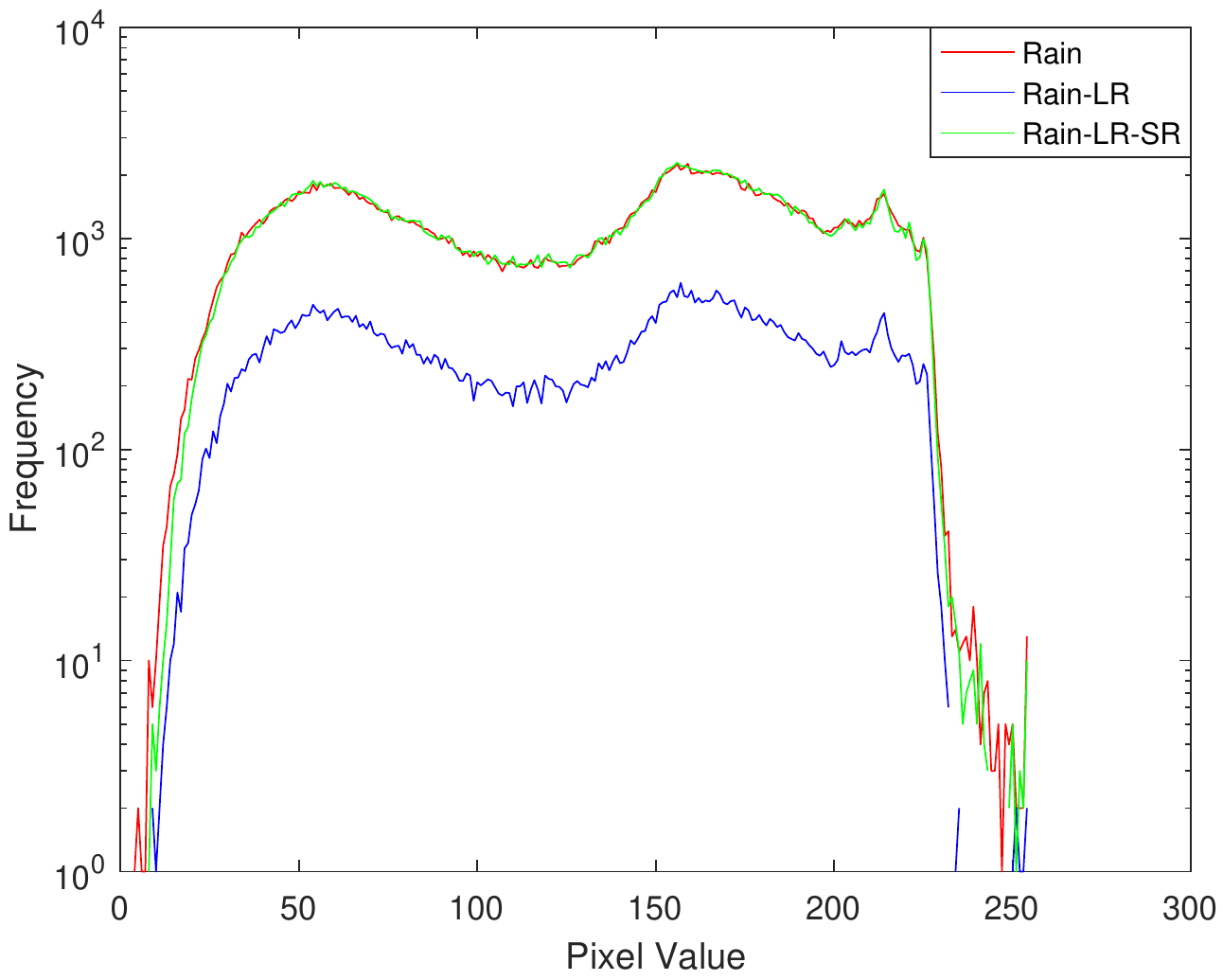} &
			\includegraphics[width=0.45\columnwidth]{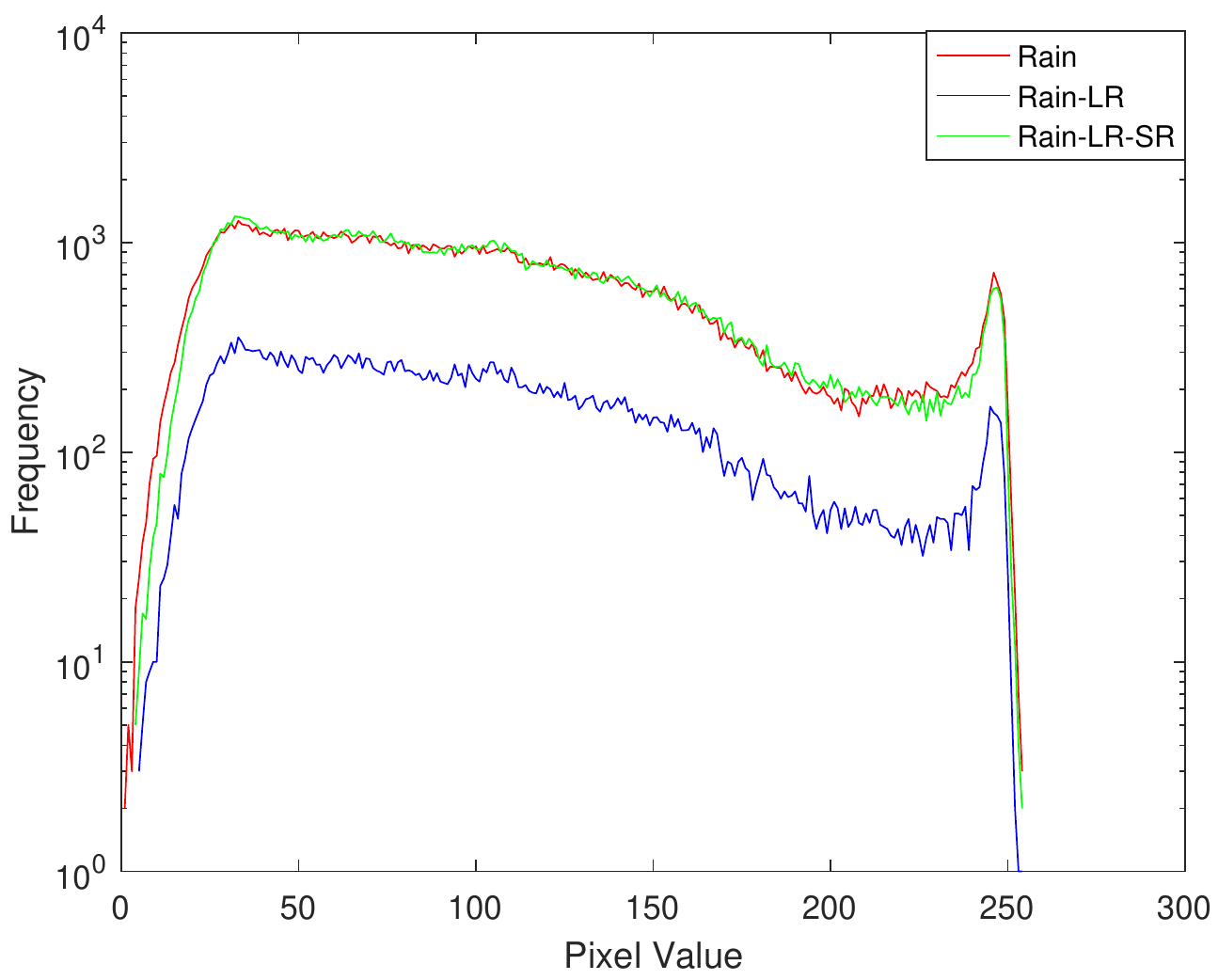} \\
            \tiny{Real-world Sample} & \tiny{Synthetic Sample}  \\
	\end{tabular}\vspace{-2mm}
\caption{Fitting results of ``Y" channel histogram for \textbf{Real} and \textbf{Synthetic} samples.
``Rain" and ``Rain-LR" denote the original and corresponding low-dimension space distribution of rainy image. ``Rain-LR-SR" is the distribution via Bilinear interpolation from ``Rain-LR". The fitting results show that the reconstructed sampling space (``Rain-LR-SR") from ``Rain-LR" can get the similar statistical distribution with that of the original space.
}
\label{fig:histogram} \vspace{-2mm}
\end{figure}

 %Section~\ref{sec:DFN}
Given a rainy image $I_{Rain}\in \mathbb{R}^{H\times W\times 3}$ and its clean version $I_{B}\in \mathbb{B}^{H\times W\times 3}$, where $H$ and $W$ denote the spatial height and weight, we observe that the reconstructed rainy image $I_{Rain, SR}\in \mathbb{R}^{H\times W\times 3}$ via bilinear interpolation from the sampled rainy image $I_{Rain,S}\in \mathbb{R}$ has the similar statistical distribution to the original one, shown in Figure~\ref{fig:histogram}. This inspires us to predict the rain streak distribution at sampling space to alleviate the learning and computational burden.
% and then reconstruct spatial details via super-resolution technologies.

In this way, we first sample $I_{Rain}$ and $I_{B}$ with Bilinear operation to generate the corresponding sub-samples ($I_{Rain,S}\in \mathbb{R}$ and $I_{B,S}\in \mathbb{B}$). As illustrated before, our ELF contains two subnetworks (IDN and BRN) to complete the image deraining via association learning. Thus, $I_{Rain,S}$ is then input to IDN to generate the corresponding rain distribution $I^*_{R,S}$ and deraining result $I^*_{B,S}$, expressed as
%predict the sub-distribution of rain streaks and produce the corresponding deraining result. The above procedures can be expressed as
\begin{equation}
\label{eq:IDN}
I^*_{R,S} = \mathcal{G}_{IDN}(F_{BS}(I_{Rain})), %I^*_{B,S} = I_{Rain,S} - I^*_{R,S},
\end{equation}
where $F_{BS}(\cdot)$ denotes the Bilinear downsampling to generate the sampled rainy image $I_{Rain,S}$. $\mathcal{G}_{IDN}(\cdot)$ refers to the rain estimation function of IDN.
%We further generate the corresponding rain-free image $I^*_{B,S}$ via by subtracting the rain residue $I^*_{R,S}$ from $I_{Rain,S}$.

Rain distribution reveals the degradation location and degree, which is naturally reasonable to be translated into the degradation prior to help accurate background recovery. Before passing $I^*_{B,S}$ into BRN for background reconstruction,
%the degradation priors are naturally reasonable to be used for accurate background recovery. To this end, we propose
a multi-input attention module (MAM),
%based on association learning,
shown in Figure~\ref{fig:Framework}, is designed to fully exploit the complementary background information from the rainy image $I_{Rain}$ via the Transformer layer, and merge them to the embedding representation of $I^*_{B,S}$. These procedures of MAM are expressed as
\begin{equation}
\label{eq:IDN1}
\begin{split}
%f_{ti} &= S(F_{K}(I^*_{R,S}) \circ F_{Q}(I_{Rain}))\circ F_{V}(I_{Rain})\\
f_{BT} &= F_{SA}(I^*_{R,S}, I_{Rain}),\\
f_{MAM} &= F_{HFB}(f_{BT}, F_{B}(I^*_{B,S})).
\end{split}
\end{equation}
In Equation~\eqref{eq:IDN1}, $F_{SA}(\cdot)$ denotes self-attention functions, involving the ebedding function and dot-product interaction. $F_{B}(\cdot)$ is the embedding function to generate the initial representation of $I^*_{B,S}$. $F_{HFB}(\cdot)$ refers to the fusion function in HFB. Following that, BRN takes $f_{MAM}$ as input for background reconstruction as
\begin{equation}
\label{eq:BRN}
I^*_{B} = \mathcal{G}_{BRN}(f_{MAM}) + F_{UP}(I^*_{B,S}),
\end{equation}
where $\mathcal{G}_{BRN}(\cdot)$ denotes the super-resolving function of BRN, and $F_{UP}(\cdot)$ is the Bilinear upsampling.

Unlike the individual training of rain streak removal and background recovery, we introduce the joint constraint to enhance the compatibility of the deraining model with background recovery, automatically learned from the training data. Then the image loss (Charbonnier penalty loss~\cite{lai2017deep,jiang2020hierarchical,hu2022spatial}) and structural similarity (SSIM)~\cite{wang2004image} loss are employed to supervise networks to achieve the image and structural fidelity restoration simultaneously. The loss functions are given by
\begin{equation}
\begin{split}
\mathcal{L}_{IDN} &= \sqrt{(I^*_{B,S}- I_{B,S})^2+\varepsilon^2} + \alpha \times SSIM(I^*_{B,S}, I_{B,S}),\\
\mathcal{L}_{BRN} &= \sqrt{(I^*_{B}- I_{B})^2+\varepsilon^2} + \alpha \times SSIM(I^*_{B}, I_{B}),\\
\mathcal{L} &= \mathcal{L}_{IDN} + \lambda \times \mathcal{L}_{BRN},
\end{split}
\label{eq:loss1}
\end{equation}
where $\alpha$ and $\lambda$ are used to balance the loss components, and experimentally set as $-0.15$ and $1$, respectively. The penalty coefficient $\varepsilon$ is set to $10^ {-3}$.

\subsection{Hybrid Fusion Network}
\label{sec:HFN}
It is known that the self-attention mechanism is the core part of Transformer, which is good at learning long-range semantic dependencies and capturing global structure representation in the image. Conversely, CNNs are skilled at modeling the local relations due to the intrinsic local connectivity. To this end, we construct the backbone of IDN and BRN into a deep dual-path hybrid fusion network by unifying the advantages of Transformer and CNNs. As shown in Figure~\ref{fig:Framework}, the backbone involves a residual Transformer branch (RTB) and an encoder-decoder branch (EDB). RTB takes a few learnable tokens (feature channels) as input and stacks multi-head attention and feed-forward networks to encode the global structure. However, capturing long-range pixel interactions is the culprit for the enormous amount of Transformer computational, making it infeasible to apply to high-resolution images, especially for the image restoration task. Besides processing the feature representation on the sampled space, inspired by~\cite{ali2021xcit}, instead of learning the global spatial similarity, we apply SA to compute cross-covariance across channels to generate the attention map encoding the global context implicitly. It has linear complexity rather than quadratically complexity.

EDB is designed to infer locally-enriched textures. Inspired by U-Net~\cite{ronneberger2015u}, we also construct EDB with the U-shaped framework. The first three stages form the encoder, and the remaining three stages represent the decoder. Each stage takes a similar architecture, consisting of sampling layers, residual channel attention blocks (RCABs)~\cite{zhang2018residual} and hybrid fusion block. Instead of using the strided or transposed convolution for rescaling spatial resolution of features, we use Bilinear sampling followed by a $1\times 1$ convolution layer to reduce checkerboard artifacts and model parameters. To facilitate residual feature fusion at different stages or scales, we design HFB to aggregate multiple inputs among stages in terms of the spatial and channel dimensions. HFB enables more diverse features to be fully used during the restoration process.

Moreover, to further reduce the number of parameters, RTB and EDB are equipped with depth-wise separable convolutions (DSC). For RTB, we integrate DSC into multi-head attention to emphasize on the local context before computing feature covariance to produce the global attention map. Moreover, we construct EDB into an asymmetric U-shaped structure, in which the encoder has the portable design with DSC, but the standard convolutions for the decoder. This scheme can save about 8\% parameters of the whole network. We have experimentally verified that utilizing DSC in the encoder is better than that in the decoder.

\begin{figure}[!t]
	\centering
	\renewcommand\arraystretch{0.3}%0.3
	\setlength\tabcolsep{0.35pt}
		\begin{tabular}{ccccc}
			\includegraphics[width=0.19\linewidth]{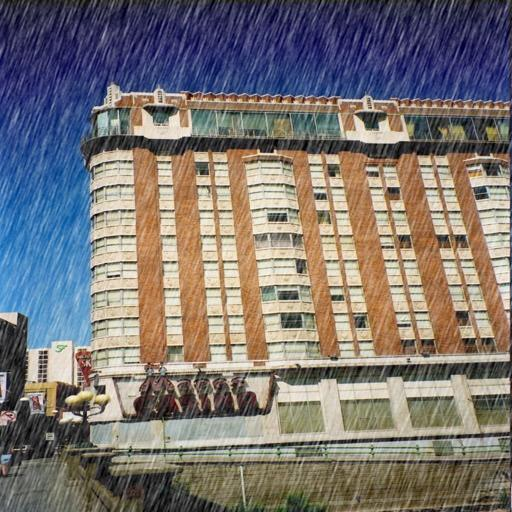} &
			\includegraphics[width=0.19\linewidth]{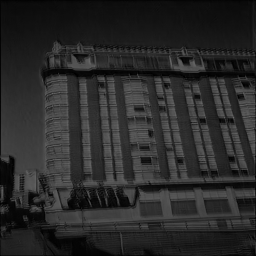} &
			\includegraphics[width=0.19\linewidth]{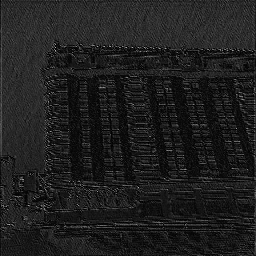} &
			\includegraphics[width=0.19\linewidth]{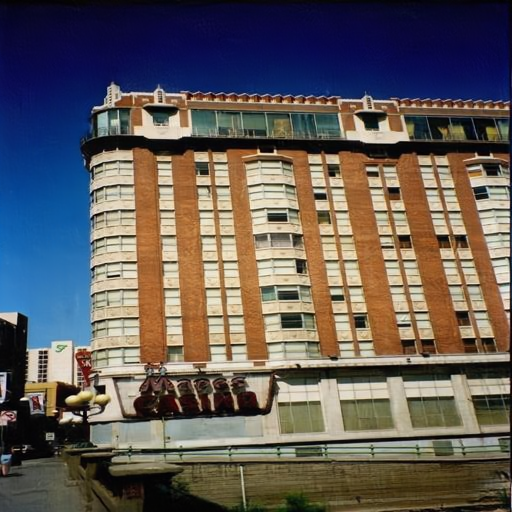} &
			\includegraphics[width=0.19\linewidth]{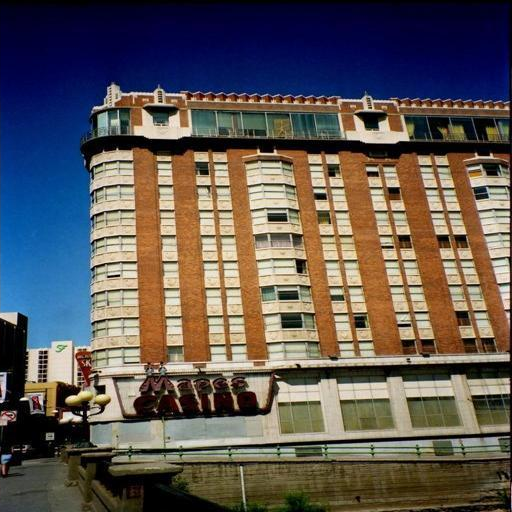} \\
			\includegraphics[width=0.19\linewidth]{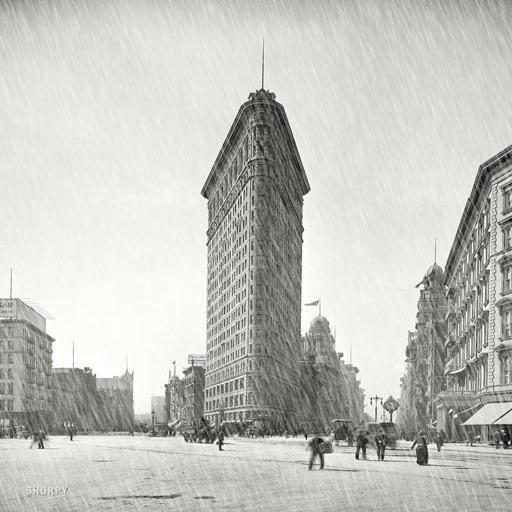} &
			\includegraphics[width=0.19\linewidth]{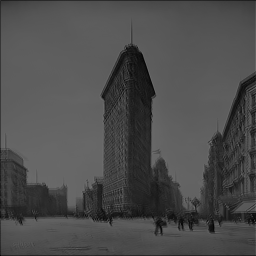} &
			\includegraphics[width=0.19\linewidth]{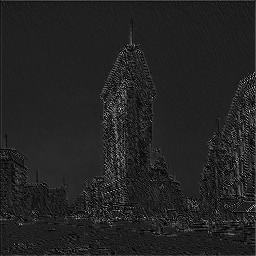} &
			\includegraphics[width=0.19\linewidth]{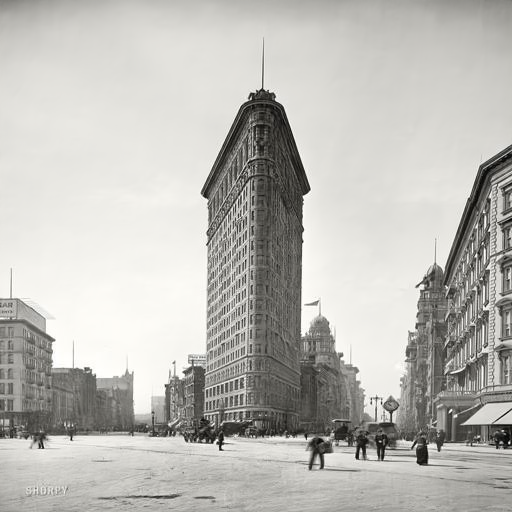} &
			\includegraphics[width=0.19\linewidth]{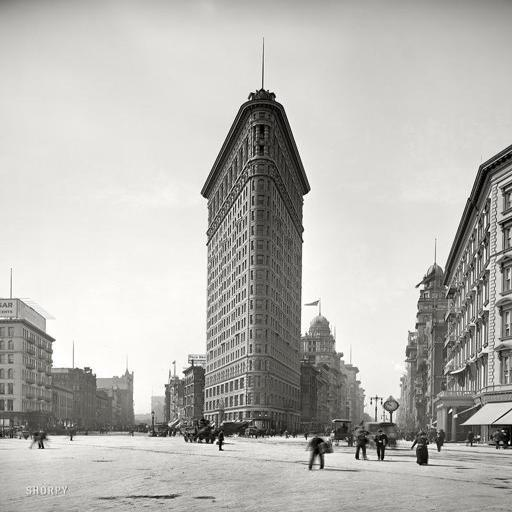}\\
			Input & $f_{B,S}$ & $f_{BT}$ & Prediction & \small{Ground Truth}\\
	\end{tabular}
	%\vspace*{-2mm}
\caption{Visualization of MAM, including the embedding representation $f_{B,S}$ of background image and the extracted background texture information ($f_{BT}$). Using the %guidance of the predicted rain distribution, the network can extract the
complementary texture $f_{BT}$ from the rainy image, the network can achieve more accurate background restoration.
%The mask provides the distortion-localization information while assigning dynamically predicted weights to the input features of undergo little or no distortion locations, helping the background restoration via the extracted complementary texture $f_{BT}$.
For a better visual effect, we respectively select three of the channels (48) from $f_{B,S}$ and $f_{BT}$, and then rescale their pixel values into [0, 255] to generate the corresponding grayscale image.}
\label{fig:MAM}\vspace*{-4mm}
\end{figure}

\subsection{Multi-input Attention Module}
\label{sec:MAM}
To associate rain streaks removal and background recovery, as shown in Figure~\ref{fig:Framework}, we construct a multi-input attention module (MAM) with Transformer to fully exploit the complementary background information for enhancement. Unlike the standard Transformer receiving a sequence of image patches as input, MAM takes the predicted rain distribution $I^*_{R,S}$, sub-space deraining image ($I^*_{B,S}$) and rainy image $I_{Rain}$ as inputs, and first learns the embedding representation ($f^*_{B,S}$, $f^*_{R,S}$, $f_{Rain}$), enriched with local contexts. $f^*_{R,S}$ and $f_{Rain}$ serve as \textit{query} (\textbf{Q}), \textit{key} (\textbf{K}) and \textit{value} (\textbf{V}) projections. Instead of learning the spatial attention map of size $\mathbb{R}^{HW\times HW}$, we then reshape query and key projections, and generate cross-covariance transposed-attention map $M \in \mathbb{R}^{C\times C}$ via the dot-product interaction between  $f^*_{R,S}$ and $f_{Rain}$. As shown in Figure~\ref{fig:MAM}, the attention map guides the network to excavate background texture information ($f_{BT}$) from the embedding representation ($f_{Rain}$) of $I_{Rain}$. The procedures in SA are expressed as
\begin{equation}
\label{eq:MAM}
%\begin{split}
%f_{MAM} &= F_{HFB}(F_{SA}(I^*_{R,S}, I_{Rain}), ,\\
F_{SA} =(Softmax(F_{K}(I^*_{R,S}) \circ F_{Q}(I_{Rain})))\circ F_{V}(I_{Rain}),
%\end{split}
\end{equation}
where $F_{K}(\cdot)$, $F_{Q}(\cdot)$ and $F_{V}(\cdot)$ are the embedding functions to produce the projections. $\circ$ and $F_{S}(\cdot)$ denote the dot-product interaction and softmax function. Followed by a hybrid fusion block, the extracted complementary information is merged with the embedding presentation of $I^*_{B,S}$ to enrich background representation.

\subsection{Hybrid Fusion Block}
\label{sec:HFB}
%The idea of residual skip connection has been widely used in CNNs to fully utilize low/high-level features. The commonly used ways are the pixel-wise superimposition and convolution-based fusion. The former is simple and fast but suffers from the redundant features, and the latter can not fully fuse the concatenated input in terms of the spatial and channel dimensions simultaneously. %However, adopting the large size convolution ($3\times 3$) requires more calculation and memory consumption while the small size one ($1\times 1$) only exploits the channel-wise dependencies between features.
Considering the feature redundancy and knowledge discrepancy among residual blocks and encoding stages, we introduce a novel hybrid fusion block (HFB) where the low-level contextualized features of earlier stages help consolidate high-level features of the later stages (or scales). Specifically, we incorporate depth-wise separable convolutions and the channel attention layer into HFB to discriminatively aggregate multi-scale features in spatial and channel dimensions. Compared to skip pixel-wise superimposition or convolution fusion, our HFB is more flexible and effective.
%Formally, taking the stage 4 of decoding part in Figure~\ref{fig:Framework} as an example, HFB takes the outputs of all encoding stages as inputs to fully exploit the multi-scale features, which are crucial for the refined restoration of degraded regions. In particular, our proposed HFB is more flexible and effective with only a few more parameters than the commonly used skip pixel-wise superimposition or convolution fusion. More analyses and discussions are presented in \textbf{Section~\ref{sec:AS}: Ablation Study}.
%Section~\ref{sec:AS}.
%\begin{equation}
%f_{S2FB} &= F_{CAL}(F_{DSC}(X_{in}, X_{out})) + F_{DSC}(X_{in}, X_{out}).
%\label{eq:S2FB}
%\end{equation}
%In Equation~\eqref{eq:S2FB}, $F_{DSC}(\cdot)$ denote the depth-wise separable convolution, composed of a depth-wise convolution and a point-wise convolution to fuse the concatenated features $X_{cat}$. $F_{CAL}(\codt)$ refers to the channel attention layer to adjust the number of channels. Taking the concatenated features as input, S2FB aggregates multi-input features from diverse perspectives, which is crucial for the refined restoration. Thanks to the effective depth-wise separable strategy, our S2FB show improved performance with only $3k$ more parameters than the superimposition fusion when the input and output channels are set to 64 and 32, respectively.
\setlength{\tabcolsep}{1.5pt}
\begin{table}[!t]
%\begin{center}
\caption{Ablation study on the depth-wise separable convolutions (DSC), multi-input attention module (MAM), hybrid fusion block (HFB), SSIM loss, super-resolution (SR), residual Transformer branch (RTB) and encoder-decoder branch (EDB) on \textit{Test1200} dataset. We obtain the model parameters (Million (M)), average inference time (Second (S)), and calculation complexity (GFlops (G)) of deraining on images with the size of $512\times 512$.}
\vspace*{-2mm}
\resizebox{1\linewidth}{!}{
\begin{tabular}{lccccccccccc}
%\hline\noalign{\smallskip}
\toprule
Model & SA & SR & DSC & HFB & MAM & SSIM & PSNR &SSIM & Par. & Time& GFlops \\%\#
%\noalign{\smallskip}
\midrule
%\noalign{\smallskip}
  Rain Image & -- & -- & --& -- & --& -- & 22.16 & 0.732 & --& --& --\\
  %\textit{w/o}~DSC & $\times$ &$\checkmark$ & $\checkmark$ & $\checkmark$& 32.57& 0.915& 1.896& 0.102& 51.70\\%51.70\\%72.38
  \textit{w/o}~SA &$\times$& $\checkmark$ & $\checkmark$&$\checkmark$ & $\checkmark$ & $\checkmark$& 32.78& 0.919& 1.536& 50.24& 50.24\\
  \textit{w/o}~DSC& $\checkmark$ & $\checkmark$ & $\times$&$\checkmark$ & $\checkmark$ & $\checkmark$& 32.73 &0.918 & 1.518& 0.121& 62.37\\
  %\textit{w/o}~HFB & $\checkmark$ &$\times$ & $\checkmark$ & $\checkmark$& 32.09& 0.914 &1.297& 0.072& 46.67\\
  \textit{w/o}~HFB& $\checkmark$ & $\checkmark$ & $\checkmark$ &$\times$ & $\checkmark$ & $\checkmark$& 32.56 &0.917 & 1.539& 0.102& 69.41\\
  \textit{w/o}~MAM& $\checkmark$ & $\checkmark$ & $\checkmark$& $\checkmark$ &$\times$&$\checkmark$ &31.46 &0.906 & 1.516& 0.121& 64.28\\
  \textit{w/o}~SSIM& $\checkmark$ & $\checkmark$ & $\checkmark$& $\checkmark$ &$\checkmark$& $\times$ & 33.17 &0.919 & 1.532& 0.125& 66.39\\
  \textit{w/o}~all& $\times$ & $\checkmark$ & $\times$ & $\times$ & $\times$&$\times$  & 29.05& 0.861 &1.538& 0.046& 61.24\\
  %\textit{w/o}~ORB & $\times$ & $\times$ & $\times$&$\times$  & 31.42& 0.896 &1.557& 0.105& 27.59\\
  %\textit{w/o}~EDB & $\times$ & $\times$ & $\times$&$\times$  & --& -- &1.583& 0.045& 101.48\\
  ELF$^*$& $\checkmark$ & $\times$ & $\checkmark$ & $\checkmark$ & $\times$& $\checkmark$ &32.97 &0.921& 1.519& 0.156& 214.67\\
  ELF& $\checkmark$ & $\checkmark$ & $\checkmark$ & $\checkmark$ &$\checkmark$& $\checkmark$ &33.38 &0.925 & 1.532& 0.125& 66.39\\
\midrule
  Model & \multicolumn{3}{c} {RTB} & \multicolumn{3}{c} {EDB}  & PSNR &SSIM &Par. & Time. & GFlops\\%\#
\midrule
  \textit{w/o}~RTB & \multicolumn{3}{c} {$\times$} & \multicolumn{3}{c} {$\checkmark$}  & 31.29& 0.910&1.536& 0.041& 26.48\\
  \textit{w/o}~EDB & \multicolumn{3}{c} {$\checkmark$} & \multicolumn{3}{c} {$\times$}   & 33.04& 0.922 &1.534& 0.154& 134.27\\
  ELF & \multicolumn{3}{c} {$\checkmark$} & \multicolumn{3}{c} {$\checkmark$} &33.38 &0.925 & 1.532& 0.125& 66.39\\
\bottomrule
\end{tabular}} \vspace{-4mm}
\label{table:ablation1}
%\end{center}
\end{table}

\section{Experiments}
\label{sec:EX}
To validate our proposed ELF, we conduct extensive experiments on synthetic and real-world rainy datasets, and compare ELF with several mainstream image deraining methods. These methods include MPRNet~\cite{zamir2021multi},  SWAL~\cite{huang2021selective}, RCDNet~\cite{wang2020model}, DRDNet~\cite{9156280}, MSPFN~\cite{9157472}, IADN~\cite{9294056}, PreNet~\cite{ren2019progressive}, UMRL~\cite{yasarla2019uncertainty}, DIDMDN~\cite{zhang2018density}, RESCAN~\cite{li2018recurrent} %,  %LPNet~\cite{fu2019lightweight}
and DDC~\cite{li2019}. Five commonly used evaluation metrics, such as Peak Signal to Noise Ratio (PSNR), Structural Similarity (SSIM), Feature Similarity (FSIM), Naturalness Image Quality Evaluator (NIQE)~\cite{mittal2012making} and Spatial-Spectral Entropy-based Quality (SSEQ)~\cite{liu2014no}, are employed for comparison.
%For a comprehensive evaluation, we also use downstream vision tasks (object detection and instance segmentation) to further validate the efficacy of our deraining approach. Moreover, due to the versatility of our proposed feature representation in low-level vision tasks, we also extend our model to other low-level image restoration tasks, such as image dehazing and low-light enhancement, for further evaluation.

\begin{table*}[!t]
  \centering
  \tiny
  \caption{Comparison results of average PSNR, SSIM, and  FSIM on Test100/Test1200/R100H/R100L datasets. When averaged across all four datasets, our ELF \textcolor{blue}{advances state-of-the-art (MPRNet) by 0.25 dB, but accounts for only its 11.7\% and 42.1\% computational cost and parameters.} We obtain the model parameters (Million) and average inference time (Second) of deraining on images with the size of \textbf{512$\times$ 512}. $^\star$ denotes the recursive network using the parameter sharing strategy. ELF-LW denotes the light-weight version of our ELF with the number of Transformer blocks to 5 and channels to 32.}
  \vspace*{-2mm}
  \resizebox{1\linewidth}{!}{
  \begin{tabular}{lccccccccccc}%GMM~\cite{li2016rain}
  \toprule
  Methods & RESCAN$^\star$~\cite{li2018recurrent}
  %& DIDMDN~\cite{zhang2018density}
  & UMRL~\cite{yasarla2019uncertainty}
  & PreNet$^\star$~\cite{ren2019progressive}
  %& LPNet %~\cite{fu2019lightweight}
  %& DDC~\cite{li2019}
  & IADN~\cite{9294056}
  & MSPFN~\cite{9157472}
  & DRDNet~\cite{9156280}
  & PCNet~\cite{jiang2021rain}
  & MPRNet~\cite{zamir2021multi}
  & SWAL~\cite{huang2021selective}
  &ELF-LW (Ours)& ELF (Ours)\\
  \midrule
  Datasets&\multicolumn{11}{c}{\textbf{Test100/Test1200}}\\
  %\cline{1-7}
  %\midrule
  PSNR &  25.00/30.51  & 24.41/30.55 & 24.81/31.36&  26.71/32.29& 27.50/32.39& 28.06/26.73& 26.17/32.03& 30.27/32.91& 28.47/30.40 & 29.41/32.55& \textbf{30.45/33.38}\\%29.90
  %\hline \hline
  SSIM  & 0.835/0.882 & 0.829/0.910 & 0.851/0.911 &  0.865/0.916& 0.876/0.916& 0.874/0.824& 0.871/0.913& 0.897/0.916& 0.889/0.892 & 0.894/0.912& \textbf{0.909/0.925}\\%0.893
  %\hline \hline
  FSIM  & 0.909/0.944 & 0.910/0.955 & 0.916/0.955 & 0.924/0.958& 0.928/0.960& 0.925/0.920& 0.924/0.956& 0.939/0.960& 0.936/0.950& 0.937/0.960& \textbf{0.945/0.964}\\%0.938
  %\hline \hline
    \midrule
  Datasets& \multicolumn{11}{c}{\textbf{R100H/R100L}}\\
  %\cline{2-7}
  %\hline \hline
  PSNR  & 26.36/29.80 & 26.01/29.18& 26.77/32.44& 27.86/32.53& 28.66/32.40& 21.21/29.24& 28.45/34.42& 30.41/36.40&29.30/34.60 & 28.83/34.61& \textbf{30.48/36.67}\\%& 26.68/36.71
  %\hline \hline
  SSIM  &0.786/0.881 & 0.832/0.923 & 0.858/0.950& 0.835/0.934& 0.860/0.933& 0.668/0.883& 0.871/0.952& 0.889/0.965&0.887/0.958 & 0.876/0.958& \textbf{0.896/0.968}\\%& 0.834/0.973
  %\hline \hline
  FSIM & 0.864/0.919 & 0.876/0.940 &0.890/0.956 & 0.875/0.942& 0.890/0.943& 0.797/0.903& 0.897/0.959& 0.910/0.969&0.908/0.963 & 0.901/0.962& \textbf{0.915/0.972}\\%&
  \midrule
  Avg-PSNR$\uparrow$  & 27.91 & 27.53& 28.84& 29.84& 30.23& 26.31& 30.27& 32.49 &30.69 & 31.35 & \textbf{32.74}\\%& 26.68/36.71
  \midrule
  Par.(M)$\downarrow$ & 0.150 & 0.984 & 0.169& 0.980& 13.35& 5.230& 0.655& 3.637& 156.54& \textbf{0.566}& 1.532 \\%2.943
  %\hline \hline
  Time (S)$\downarrow$ & 0.546 & 0.112 & 0.163& 0.132& 0.507& 1.426& \textbf{0.062}& 0.207&0.116& 0.072& 0.125\\%0.110
  GFlops (G)$\downarrow$ & 129.28 & 65.74 & 265.76& 80.99& 708.39& --& 28.21& 565.81& 614.35& \textbf{21.53}& 66.39\\%0.110
  \bottomrule
  \end{tabular}}\vspace{-2mm}
  \label{table:synthetic1}
\end{table*}

\subsection{Implementation Details}
\label{sec:ID}
\textbf{Data Collection.}
Since there exits the discrepancy in training samples for all comparison methods, following~\cite{9157472,9294056}, we use $13,700$ clean/rain image pairs from~\cite{zhang2019image,fu2017removing} for training all comparison methods with their publicly released codes by tuning the optimal settings for a fair comparison.
%These samples cover various rain streak orientations and magnitudes under distinct rain conditions.
%For fairness, the competing methods are retrained in the experiments with publicly released codes following their original settings on this common training dataset.
For testing, four synthetic benchmarks (Test100~\cite{zhang2019image}, Test1200~\cite{zhang2018density}, R100H and R100L~\cite{yang2017deep}) and three real-world datasets (Rain in Driving (RID), Rain in Surveillance (RIS)~\cite{li2019single} and Real127~\cite{zhang2018density}) are considered for evaluation.

\noindent \textbf{Experimental Setup.}
\label{sec:ex}
In our baseline, the number of Transformer blocks in RTB is set to 10 while RCAB is empirically set to $1$ for each stage in EDB with filter numbers of 48. The training images are coarsely cropped into small patches with a fixed size of $256\times 256$ pixels to obtain the training samples. We use Adam optimizer with the learning rate ($2\times10^{-4}$ with the decay rate of 0.8 at every 65 epochs till 600 epochs) and batch size ($12$) to train our ELF on a single Titan Xp GPU.
%  with the decay rate of 0.8 at every $100,000$ steps till $1\times10^{-6}$. We train the network for $60$ epochs with the above settings.

\subsection{Ablation Study}
\label{sec:AS}

\textbf{Validation on Basic Components}.
We conduct ablation studies to validate the contributions of individual components, including the self-attention (SA), depth-wise separable convolutions (DSC), super-resolution reconstruction (SR), hybrid fusion block (HFB) and multi-input attention module (MAM) to the final deraining performance. For simplicity, we denote our final model as ELF and devise the baseline model \textit{w/o}~all by removing all these components above. Quantitative results in terms of deraining performance and inference efficiency on the Test1200 dataset are presented in Table~{\ref{table:ablation1}}, revealing that the complete deraining model ELF achieves significant improvements over its incomplete versions. Compared to \textit{w/o~MAM} model (removing MAM from ELF), ELF achieves 1.92dB performance gain since the association learning in MAM can help the network to fully exploit the background information from the rainy input with the predicted rain distribution prior. In addition, disentangling the image deraining task into rain streaks removal and texture reconstruction at the low-dimension space exhibits considerable superiority in terms of efficiency (19.8\% and 67.6\% more efficient in inference time and computational cost, respectively) and restoration quality (referring to the results of ELF and ELF$^*$ models (accomplishing the deraining and texture recovery at the original resolution space~\cite{9156280})). Moreover, using the depth-wise separable convolution allows increasing the channel depth with approximately the same parameters, thus enhancing the representation capacity (referring to the results of ELF and \textit{w/o~DSC} models). Compared to the \textit{w/o~SA} that replaces the Transformer blocks in RTB with the standard RCABs, ELF gains 0.45dB improvement with acceptable computational cost.

We also conduct ablation studies to validate the dual-path hybrid fusion framework, involving an residual Transformer branch (RTB) and a U-shaped encoder-decoder branch (EDB). Based on ELF, we devise two comparison models (\textit{w/o}~RTB and \textit{w/o}~EDB) by removing these two branches in turn. Quantitative results are presented in Table~{\ref{table:ablation1}}. Removing RTB may greatly weaken the representation capability on the spatial structure, leading to the obvious performance decline (2.09dB in PSNR) (referring to the results of ELF and \textit{w/o}~RTB models). Moreover, EDB allows the network to aggregate multi-scale textural features, which is crucial to enrich the representation of local textures.
%for fine detail recovery.

\begin{figure*}[!ht]
\flushleft
\centering
\includegraphics[width=0.98\textwidth]{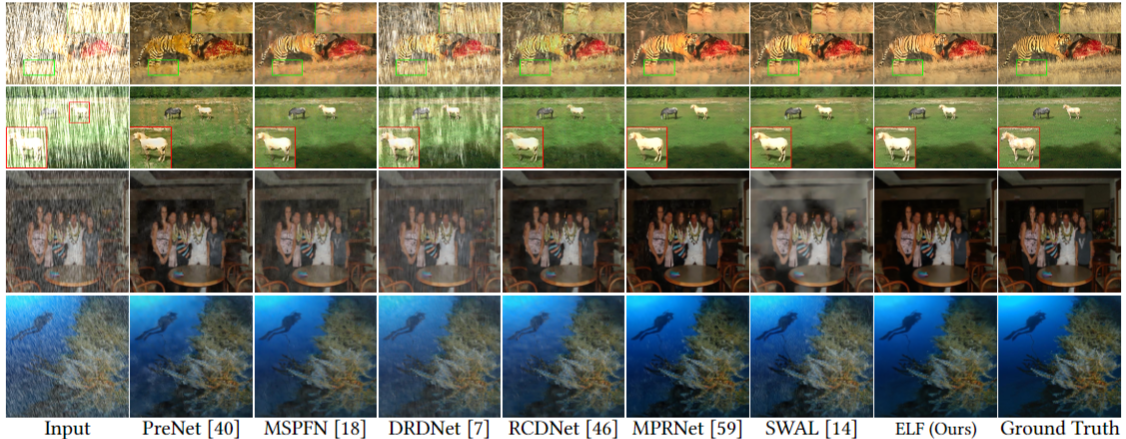}
\vspace*{-4mm}%%\vspace*{-4mm}
\caption{Visual comparison of derained images obtained by seven methods on R100H/R100L ($1^{st}-2^{th}$ rows) and Test100/Test1200 ($3^{st}-4^{th}$ rows) datasets. Please refer to the region highlighted in the boxes for a close up comparison.}
\label{fig:synthetic1}
\vspace*{-4mm}
\end{figure*}

\subsection{Comparison with State-of-the-arts}
\label{sec:CS}
\textbf{Synthesized Data.}
Quantitative results on Test1200, Test100, 100H and R100L datasets are provided in Table~{\ref{table:synthetic1}}. Meanwhile, the inference time, model parameters and computational cost are also compared. It is observed that most of the deraining models obtain impressive performance on light rain cases with high consistency. However, only our ELF and MPRNet still perform favorably on heavy rain conditions, exhibiting great superiority over other competing methods in terms of PSNR. As expected, our ELF model achieves the best scores on all metrics, \textcolor{blue}{surpassing the CNN-based SOTA (MPRNet) by 0.25 dB on average, but only accounts for its 11.7\% and 42.1\% computational cost and parameters.} Meanwhile, our light-weight deraining model ELF-LW is still competitive, which gains the third-best average PSNR score on four datasets. In particular, ELF-LW averagely surpasses the real-time image deraining method PCNet~\cite{jiang2021rain} by 1.08dB, while with less parameters (saving  13.6\%) and computational cost (saving 23.7\%).
%The results on R100H/R100L datasets are provided in Table~{\ref{table:synthetic1}}, indicating that most of the deraining models obtain impressive performance on light rain cases with high consistency. As expected, our ELF model achieves the best scores on all metrics, surpassing the CNN-based SOTA (MPRNet) by 0.25 dB on average, but only accounts for its 11.7\% and 42.1\% computational cost and parameters.
%much better restoration performance than these competitors on almost all evaluation metrics, surpassing MSPFN~\cite{9157472} and DRDNet~\cite{9156280} in PSNR by 2.40dB and 1.94dB respectively on Test100 dataset, while it is 78.6\% and 78\% more efficient in inference time and parameters than MSPFN.
%Compared with the light-weight deraining method, such as LPNet~\cite{fu2019lightweight}, ELF is still very competitive in both recovery quality and inference time. Although LPNet has faster inference time and less parameters, its deraining performance is up to 5.84dB and 7.57dB (PSNR) less than that of our ELF on Test100 and Test1200 datasets.
\begin{table*}[!t]
\begin{center}
%\tiny
\caption{Comparison of average NIQE/SSEQ scores with ten deraining methods on three real-world datasets.}\vspace*{-2mm}
\resizebox{1\linewidth}{!}{
\begin{tabular}{lccccccccccc}
\toprule
%hline\noalign{\smallskip}%Datasets(Num)
 Datasets & DIDMDN %~\cite{zhang2018density}
 & RESCAN~\cite{li2018recurrent}
 & DDC~\cite{li2019}
 & LPNet~\cite{fu2019lightweight}
 & UMRL~\cite{yasarla2019uncertainty}
 & PreNet~\cite{ren2019progressive}
 & IADN~\cite{9294056}
 & MSPFN~\cite{9157472}
 & DRDNet~\cite{9156280}
 & MPRNet~\cite{zamir2021multi}
 & ELF (Ours)\\%& PreNet
\noalign{\smallskip}
\midrule
\noalign{\smallskip}
  Real127 (127) %~\cite{zhang2018density}
  & 3.929/32.42 & 3.852/30.09 & 4.022/29.33 & 3.989/29.62 & 3.984/29.48& 3.835/29.61& 3.769/29.12 & 3.816/\textbf{29.05}& 4.208/30.34&3.965/30.05& \textbf{3.735}/29.16\\%
  RID (2495) %~\cite{li2019single}
  & 5.693/41.71 & 6.641/40.62 & 6.247/40.25 & 6.783/42.06 & 6.757/41.04 & 7.007/43.04& 6.035/40.72 & 6.518/40.47& 5.715/39.98&6.452/40.16& \textbf{4.318/37.89}\\
  RIS (2348) %~\cite{li2019single}
  & 5.751/46.63 & 6.485/50.89 & 5.826/47.80 & 6.396/53.09 & \textbf{5.615}/43.45 & 6.722/48.22& 5.909/42.95& 6.135/43.47& 6.269/45.34&6.610/48.78& 5.835/\textbf{42.16}\\
\bottomrule
\end{tabular}}
\label{table:real}
\vspace*{-4mm}
\end{center}
\end{table*}

For more convincing evidence, we also provide visual comparisons in Figure~\ref{fig:synthetic1}. High-accuracy methods, such as PreNet, MSPFN and RCDNet, can effectively eliminate the rain layer and thus bring an improvement in visibility. But they fail to generate visual appealing results \textit{by introducing considerable artifacts and unnatural color appearance}, the heavy rain condition in particular. Likewise, DRDNet focuses on the detail recovery, but shows undesired deraining performance. MPRNet tends to produce over-smoothing results. Besides recovering cleaner and more credible image textures, our ELF produces results with better contrast and less color distortion. Please refer to the ``tiger" and ``horse" scenarios. Moreover, we provide the comparison and analyses in terms of the color histogram fitting curve of ``Y" channel in \textcolor{blue}{Supplementary}, which verifies the consistency between the predicted deraining result to the ground truth in terms of the statistic distribution.
We speculate that these visible improvements on restoration quality may benefit from our proposed hybrid representation framework of Transformer and CNN as well as the association learning scheme for rain streak removal and background recovery. These strategies are integrated into a unified framework, allowing the network to fully exploit the respective learning merits for image deraining while guaranteeing the inference efficiency.

\begin{figure*}[!ht]
\flushleft
\centering
\includegraphics[width=0.98\textwidth]{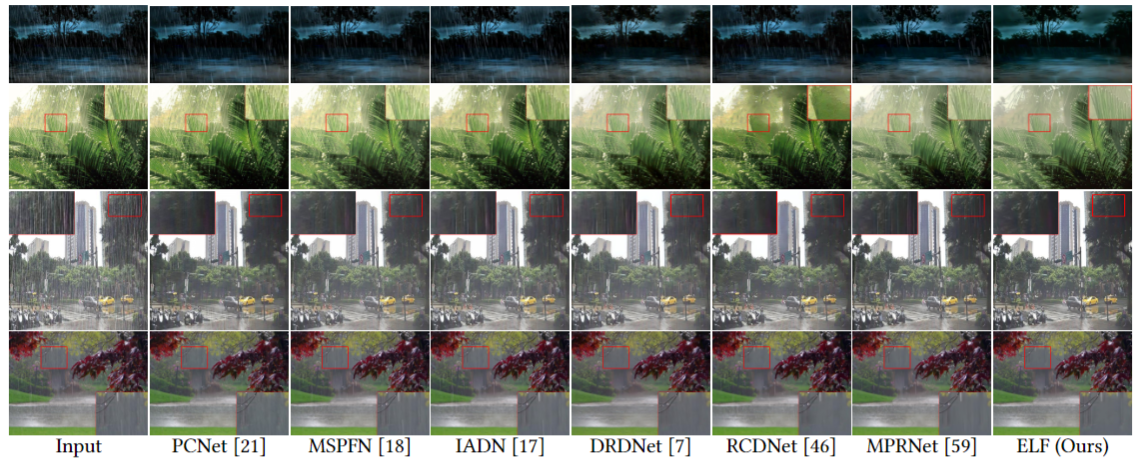}
	  \vspace*{-2mm}
\caption{Visual comparison of derained images obtained by eight methods on five \textcolor{red}{real-world scenarios}, coverring rain veiling effect (1$^{st}$), heavy rain (2$^{st}$) and light rain (3$^{st}$-4$^{st}$). Please refer to the region highlighted in the boxes for a close up comparison.}
\label{fig:real}
  \vspace*{-2mm}
\end{figure*}

\begin{figure*}[!ht]
\flushleft
\centering
\includegraphics[width=0.98\textwidth]{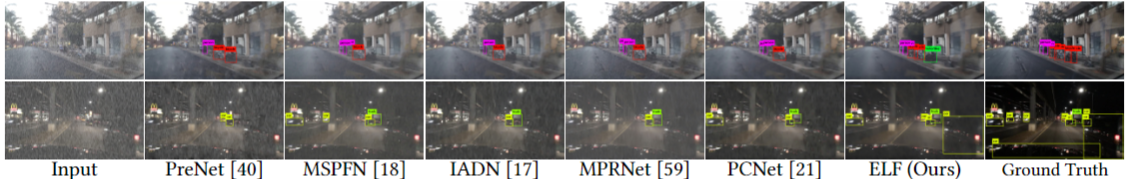}
	  \vspace*{-2mm}
\caption{Visual comparison of joint image deraining and object detection on BDD350 dataset.}
\label{fig:detection}
  %\vspace*{-2mm}
\end{figure*}

%\begin{figure*}[ht!]
%\centering
%\includegraphics[width=7in]{./img/real127/real.pdf}
%\vspace{-0.7em}
%\caption{Visual comparison of de-rained images obtained by seven  methods under four real-world %scenarios.}%\vspace{1.0em}
%\label{fig:real127}
%\end{figure*}
%\setlength{\tabcolsep}{4pt}
%\tiny%\small

\noindent \textbf{Real-world Data.}
We further conduct experiments on three real-world datasets: Real127~\cite{zhang2018density}, Rain in Driving (RID), and Rain in Surveillance (RIS)~\cite{li2019single}.
%RID and RIS respectively contain 2495 and 2348 real-world rain samples collected from car-mounted cameras and networked traffic surveillance cameras in rainy days. These datasets differ in rain types, image quality, object size, angle, etc, and represent real application scenarios where deraining may be desirable.
Quantitative results of NIQE~\cite{mittal2012making} and SSEQ~\cite{liu2014no} are listed in Table~{\ref{table:real}}, where smaller NIQE and SSEQ scores indicate better perceptual quality and clearer contents. Again, our proposed ELF is highly competitive, achieving the lowest average values on the RID dataset and the best best average scores of NIQE and SSEQ on the Real127 and RIS datasets, respectively. We visualize the deraining results in Figure~\ref{fig:real}, showing that ELF produces rain-free images with cleaner and more credible contents, whereas the competing methods fail to remove rain streaks. These evidences indicate that our ELF model performs well in eliminating rain perturbation while preserving textural details and image naturalness.

\begin{table*}[!t]
\begin{center}
\scriptsize
\caption{Comparison results of joint image deraining and object detection on COCO350/BDD350 datasets.}\vspace*{-4mm}
\resizebox{1\linewidth}{!}{
\begin{tabular}{lcccccccc}
\toprule
Methods & Rain input & RESCAN~\cite{li2018recurrent}
%& LPNet %~\cite{fu2019lightweight}
%& DDC %~\cite{li2019}
& PreNet~\cite{ren2019progressive}
& IADN~\cite{9294056}
& MSPFN~\cite{9157472}
&MPRNet~\cite{zamir2021multi}
&ELF-LW
& ELF (Ours)\\
\noalign{\smallskip}
\midrule
\noalign{\smallskip}
\multicolumn{9}{c} {\textcolor[rgb]{0.50,0.00,1.00}{Deraining}; Dataset: \textbf{COCO350/BDD350}; Image Size: \textbf{640$\times$ 480}/\textbf{1280$\times$ 720}}\\
%\cline{1-7}
\midrule
  %PSNR  & 14.79/14.13  &  17.04/16.71& 15.43/14.87 & 17.01/16.85& 17.53/16.90 &18.18/17.91& 18.23/17.85&\textbf{18.51/17.97}\\
  PSNR  & 14.79/14.13  &  17.04/16.71& 17.53/16.90 &18.18/17.91& 18.23/17.85&17.99/16.83&18.43/18.09&\textbf{18.93/18.49}\\
  %\hline
  SSIM & 0.648/0.470 & 0.745/0.646&  0.765/0.652 &0.790/0.719& 0.782/\textbf{0.761}&0.769/0.622&0.800/0.714&\textbf{0.818/0.761}\\
  %SSIM & 0.648/0.470 & 0.745/0.646 &0.677/0.520 & 0.748/0.662&  0.765/0.652 &0.790/0.719& 0.782/\textbf{0.761}&\textbf{0.804}/0.725\\
  %Ave.inf.time (s) & --/-- &  0.546/1.532 & 0.032/0.082 & 0.136/0.407& 0.227/0.764 & 0.135/0.412& 0.584/1.246 & 0.082/0.118\\
  Ave.inf.time (s) & --/-- &  0.546/1.532 & 0.227/0.764 & 0.135/0.412& 0.584/1.246 & 0.181/0.296&\textbf{0.076/0.160} &0.128/0.263\\
\midrule
\multicolumn{9}{c}{\textcolor[rgb]{1.00,0.00,0.00}{Object Detection}; Algorithm: \textbf{YOLOv3}; Dataset: \textbf{COCO350/BDD350}; Threshold: 0.6}\\%~\cite{redmon2018yolov3}
%\cline{1-7}
\midrule
%   Precision ($\%$) & 23.03/36.86 &  28.74/40.33  &24.87/36.59 & 29.66/39.02&  31.31/38.66 &32.92/40.28& 32.56/41.04&\textbf{33.51/41.35}\\
%   %\hline
%   Recall ($\%$) &  29.60/42.80 &  35.61/47.79  &31.45/43.74 & 37.13/48.15&  37.92/48.59 &39.83/50.25& 39.31/50.40 &\textbf{40.63/51.75}\\
%   %\hline
%   IoU ($\%$) & 55.50/59.85 &59.81/61.98 &56.58/59.55& 60.02/61.60&  60.75/61.08 &61.96/62.27& 61.69/62.42& \textbf{62.40/62.53}\\
  Precision ($\%$) & 23.03/36.86 &  28.74/40.33&  31.31/38.66 &32.92/40.28& 32.56/41.04&31.31/40.49&33.31/40.88&33.85/41.84\\
  %\hline
  Recall ($\%$) &  29.60/42.80 &  35.61/47.79&  37.92/48.59 &39.83/50.25& 39.31/50.40 &38.98/48.77&40.43/51.55&40.43/52.60\\
  %\hline
  IoU ($\%$) & 55.50/59.85 &59.81/61.98&  60.75/61.08 &61.96/62.27& 61.69/62.42& 61.14/61.99&62.21/62.48&62.61/62.80\\
\bottomrule
\end{tabular}} \vspace*{-2mm}
\label{table:dtection}
\end{center}
\end{table*}

\subsection{Impact on Downstream Vision Tasks}
\label{sec:op}
Eliminating the degraded effects of rain streaks under rainy conditions while preserving credible textural details is crucial for object detection. This motivates us to investigate the effect of deraining performance on object detection accuracy based on popular object detection algorithms (\textit{e.g.}, YOLOv3~\cite{redmon2018yolov3}). Based on our ELF and several representative deraining methods, the restoration procedures are directly applied to the rainy images to generate corresponding rain-free outputs. We then apply the publicly available pre-trained models of YOLOv3 for the detection task.
Table~{\ref{table:dtection}} shows that ELF achieves highest PSNR scores on COCO350 and BDD350 datasets~\cite{9157472}. Meanwhile, the rain-free results generated by ELF facilitate better object detection performance than other deraining methods. Visual comparisons on two instances in Figure~\ref{fig:detection} indicate that the deraining images by ELF exhibit a notable superiority in terms of image quality and detection accuracy. We attribute the considerable performance improvements of both deraining and down-stream detection tasks to our association learning between rain streaks removal and detail recovery tasks.
%, which provides more discriminative information to promote the detection accuracy.
%Visual comparison results and the corresponding detection precision on two scenarios are shown in Figure~\ref{fig:detection}. It is evident that rain streaks can greatly degrade the detection accuracy by missing targets and producing a low detection precision.
%Visual examples of the joint image deraining and object detection are shown in the \textit{Supplementary Material}.

\subsection{Robust Analyses on Adversarial Attacks}
\label{sec:RAAAA}
We conduct a brief study on the robustness of mainstream rain removal methods against adversarial attacks, including the LMSE attack and LPIPS attacks~\cite{yu2022towards}. Our study shows that the deraining models are more vulnerable to the adversarial attacks perturbations while our method shows better robustness over other competitors. More details and analyses are included in \textcolor{blue}{Supplementary}.

\section{Conclusion}
\label{sec:CO}
We rethink the image deraining as a composite task of rain streak removal, textures recovery and their association learning, and propose %an efficient and accurate enhancement model called perturbation-guidance texture reconstruction network (PTRNet) for mainstream image enhancement tasks.
a dynamic associated network (ELF) for image deraining.
%PTRNet contains two sub-networks to jointly tackle perturbation removal and high-frequency detail reconstruction.
Accordingly, a two-stage architecture and an associated learning module (ALM) are adopted in ELF to account for twin goals of rain streak removal and texture reconstruction while facilitating the learning capability. Meanwhile, the joint optimization promotes the compatibility while maintaining the model compactness.
%Moreover, we propose a perturbation-guidance texture enhancement module (ALM) to fully exploit the input information under the guidance of predicted perturbation priors.  simultaneously realize the twin goals of restoration and reconstruction
%A selective fusion block (SFB) is also designed to aggregate the multi-scale features among residual locks and encoding stages.
Extensive results on image deraining and joint detection task demonstrate the superiority of our ELF model over the state-of-the-arts. %In addition, we also find that our method works well in mainstream image enhancement tasks, \textit{i.e.}, image dehazing and low-light enhancement (see the \textcolor{blue}{Supplementary}).

%%
%% The acknowledgments section is defined using the "acks" environment
%% (and NOT an unnumbered section). This ensures the proper
%% identification of the section in the article metadata, and the
%% consistent spelling of the heading.
\begin{acks}
This work is supported by National Natural Science Foundation of China (U1903214,62071339, 62072347, 62171325), Natural Science Foundation of Hubei Province (2021CFB464)
Guangdong-Macao Joint Innovation Project (2021A0505080008)
Open Research Fund from Guangdong Laboratory of Artificial Intelligence and Digital Economy (SZ)(GML-KF-22-16).
%This work is supported by National Natural Science Foundation of China (U1903214, 62071339, 61872277, 62072347, 62171325), Natural Science Foundation of Hubei Province (2021CFB464), Open Research Fund from Guangdong Laboratory of Artificial Intelligence and Digital Economy (SZ)(GML-KF-22-16) and Guangdong-Macao Joint Innovation Project (2021A0505080008).
\end{acks}

\bibliographystyle{ACM-Reference-Format}
\bibliography{sample-base}
\end{document}